\documentclass{article}

\usepackage[nonatbib,final]{neurips_2021}

\usepackage[utf8]{inputenc} 
\usepackage[T1]{fontenc}    
\usepackage{hyperref}       
\usepackage{url}            
\usepackage{booktabs}       
\usepackage{amsfonts}       
\usepackage{nicefrac}       
\usepackage{microtype}      

\usepackage[dvipsnames,table,xcdraw]{xcolor}         
\usepackage{wrapfig,lipsum,booktabs}

\usepackage{graphicx}

\newcommand{\cmark}{\ding{51}}%
\newcommand{\xmark}{\ding{55}}%

\usepackage{algorithm} 
\usepackage{algorithmic} 
\usepackage{colortbl}

\usepackage{amssymb}
\usepackage{pifont}

\def\eg{\textit{e.g.}}
\def\ie{\textit{i.e.}}

\def\wrt{\textit{w.r.t.}}
\def\vs{\textit{v.s.}}

\def\lofi{LoFi}
\def\tlofi{T-LoFi}
\def\slofi{S-LoFi}
\def\stlofi{ST-LoFi}
\def\clofi{C-LoFi}
\def\LF{LoFi}

\definecolor{Gray}{gray}{0.85}

\title{{Low-Fidelity Video Encoder Optimization} for Temporal Action Localization}

\author{Mengmeng Xu$^{1,2}\thanks{Work done during an internship at Samsung AI Centre.}$ \\
\texttt{mengmeng.xu@kaust.edu.sa}
\And
Juan-Manuel P\'erez-R\'ua$^{1}$ \\
\texttt{PerezRua.JM@gmail.com}
\And
Xiatian Zhu$^{1}$ \\
\texttt{xiatian.zhu@samsung.com}
\And 
Bernard Ghanem$^{2}$ \\
\texttt{bernard.ghanem@kaust.edu.sa}
\And
Brais Martinez$^{1}$ \\
\texttt{brais.a@samsung.com}\\
}

\begin{document}

\maketitle
\vspace{-20pt}
\begin{center}
{\small $^1$ Samsung AI Centre Cambridge, UK}
{\small $^2$ King Abdullah University of Science and Technology, Saudi Arabia}
\end{center}
\vspace{20pt}

\begin{abstract}
{Most} {existing {temporal action localization (TAL)} methods rely on a transfer learning pipeline, first optimizing a video encoder on a large action classification dataset (\ie, source domain), followed by freezing the encoder and training a TAL head on the {action localization} dataset (\ie, target domain).} This results in a task discrepancy problem for the video encoder -- trained for action classification, but used for TAL.
%
{Intuitively, joint optimization with both the video encoder and TAL head is an obvious solution to} {this discrepancy.} However, this is not operable for TAL subject to the GPU memory constraints, due to the prohibitive computational cost in processing long untrimmed videos.
In this paper, we resolve this challenge by
introducing a novel low-fidelity (\lofi{}) video encoder optimization method.
Instead of always using the full training configurations in TAL learning, we propose to reduce the mini-batch composition in terms of 
temporal, spatial or spatio-temporal resolution
so that {jointly optimizing the video encoder and TAL head} becomes operable under the same memory conditions of a mid-range hardware budget. 
Crucially, this enables the gradients to flow backwards through the video encoder conditioned on a TAL supervision loss, favourably solving the task discrepancy problem and providing more effective feature representations.
Extensive experiments show that the proposed \lofi{} {optimization} approach can significantly enhance the performance of existing TAL methods. Encouragingly, even with a lightweight ResNet18 based video encoder in a single RGB stream, our method surpasses two-stream {(RGB + optical flow)} ResNet50 based alternatives, often by a good margin.
\end{abstract}

\section{Introduction}
Video analysis has 
become an increasingly important area of research, encompassing multiple relevant problems such as action recognition \cite{carreira2017quo,feichtenhofer2016convolutional}, temporal action localization~\cite{zhao2019hacs,CabaHeilbron2016,Escorcia2016DAPsDA,caba2015activitynet,xu2020boundary,xu2020g,ProposalRelationNetwork,zhao2021video,Gao_Shi_Wang_Li_Yuan_Ge_Zhou_2020}, video grounding\cite{liu2020jointly,zhang_etal_2020_span,Zeng_2020_CVPR,2DTAN_2020_AAAI,Soldan_2021_ICCV, gao2021relation, Huang_2021_ICCV}, and video question answering \cite{huang2020aligned,lei2018tvqa}. 
Among those, temporal action localization (TAL) 
\cite{zhao2019hacs,caba2015activitynet} is a fundamental 
task, as natural videos are not temporally trimmed.
Given an untrimmed video, TAL aims to identify the start and end points of all action instances and recognize their category labels simultaneously.
A typical TAL {model} is based on deep convolutional neural networks (CNNs) composed of two modules: 
a video encoder and a TAL head.
The video encoder is often shared across different TAL {methods} (\eg, G-TAD~\cite{xu2020g}, BC-GNN~\cite{Bai2020bcgnn}) 
by taking a specific off-the-shelf action classification model (\eg, C3D~\cite{7410867}, I3D~\cite{carreira2017quo}, TSM~\cite{lin2019tsm}), with the differences residing only in the TAL head.
However, instead of short (\eg, 10 seconds) trimmed video clips as in action recognition, {the input videos to a TAL model are} characterized by much longer temporal duration {(\eg, 120 seconds)}.
%
This causes unique computational challenges that remain
unsolved, particularly in model optimization.

In standard optimization of a TAL model, a two-stage {transfer learning pipeline} is often involved:
\begin{enumerate}
    \item First, the video encoder is optimized on a large source video classification dataset (\eg, Kinetics~\cite{zisserman2017kinetics})
    and, optionally, {finetunned} on the {trimmed} version of the target dataset
    under {\em action classification supervision};
    \item Second, {the video encoder is frozen} and the TAL head is optimized on the target action localization dataset (\eg, ActivityNet~\cite{caba2015activitynet}, {HACS~\cite{zhao2019hacs}}) under 
    {\em TAL task supervision}. 
\end{enumerate}

With this widely-used TAL training pipeline, the video encoder is only optimal for action classification but {not for 
the target TAL task.} {Specifically, the video encoder is trained so that different short segments within an action sequence are mapped to similar outputs, thus encouraging insensitivity to the temporal boundaries of actions}.
This is not desirable for a TAL model. 
We identify this as \textbf{\em a task discrepancy problem}.
Consequently, the final TAL model could suffer from suboptimal performance. 

{Indeed, jointly optimizing all components of a CNN architecture
end-to-end with the target task's supervision is a common practice, \eg, training models for object detection in static images~\cite{10.1109/CVPR.2014.81,redmon2015unified,Liu_2016}.}
Unfortunately, 
{this turns out to be non-trivial for TAL.}
As mentioned above, model training is severely restricted
by the large input size of untrimmed videos and
subject to the memory constraint of GPUs.
This is why the two-{stage optimization pipeline} as described above becomes the most common and feasible choice in practice for optimizing a TAL model.
{On the other hand, existing transfer learning methods
mostly focus on tackling the data distribution shift problem across
different datasets~\cite{9134370,10.1007/978-3-030-01424-7_27}, rather than the task shift problem we study here.}
Regardless, we believe that solving this limitation of the TAL training design bears a great potential for improving model performance. 

In this work, we present a simple yet effective 
{\em low-fidelity} (\lofi{}) video encoder optimization method
particularly designed for better TAL model training.
{It is designed to adapt the video encoder from action classification to TAL whilst subject to the same hardware budget.}
This is achieved by introducing a 
simple strategy characterized 
by a new intermediate training stage 
where 
both the video encoder and the TAL head {are optimized} end-to-end using a lower temporal and/or spatial resolution (\ie, low-fidelity) in the mini-batch construction.
Compared to the standard training method, our proposed strategy does not increase the GPU memory standard
(often a hard constraint for many practitioners).
%
%
Crucially, with our \lofi{} training
the gradients back-propagate to the video encoder from a temporal action localization loss whilst conditioned on the target TAL head, enabling the learning of a video encoder sensitive to the temporal localization objective. 

We make the following {\bf contributions} in this work.
{\bf (1)}
We investigate the limitations of the standard optimization method for TAL models, and consider that the task discrepancy problem hinders the performance of existing TAL models.
Despite it being a significant ingredient, video encoder optimization is largely ignored by existing TAL methods, left without systematic investigation.
{\bf (2)}
To improve the training of TAL models, we present a novel, simple, and effective {\em low-fidelity} (\lofi{}) video encoder optimization method.
{It is designed specifically to address the task discrepancy problem with the TAL model's video encoder.}
{\bf (3)}
Extensive experiments show that the proposed \lofi{} {optimization method} yields new state-of-the-art performance when combined with 
off-the-shelf TAL models (\eg, G-TAD \cite{xu2020g}).
Critically, our method achieves superior efficiency/accuracy trade-off
with clear inference cost advantage and good generalizability
to varying-capacity video encoders.


\section{Related Work}
\paragraph{Temporal action localization (TAL) models:}
TAL models can be grouped by architectural design pattern into two categories, one-stage and two-stage architectures.
One-stage methods, either predict temporal action boundaries or generate proposals, and classify them within the same network~\cite{Bai2020bcgnn,chao2018rethinking,heilbron2017scc, Long2019GaussianTA, lin2019bmn,xu2017r,xu2020g, Yuan2017TemporalAL}.
The latter type, two-stage models, generate sets of action proposals (\eg, segments)~\cite{Buch2017SSTST, Escorcia2016DAPsDA, Gao2018CTAPCT,Heilbron2016FastTA, Liu2019MultiGranularityGF} and then an auxiliary head is used for classification of each proposal into an action class~\cite{lin2018bsn,shou2017cdc,Shou2016TemporalAL, Zeng2019GraphCN, Zhao2017TemporalAD}. 
In this work, rather than introducing a novel model design,
we focus on the training of generic TAL models, with a particular aim to improve the video encoder optimization.
This is a relatively less investigated aspect in the TAL literature.

\paragraph{Video encoders in TAL:}
The video encoder is an indispensable part of a TAL model. 
{Main design choices include the base architecture of video encoder and its optimization procedure. With regards to the architecture,} the two-stream Temporal Segment Network (TSN)~\cite{wang2016temporal} is one of the most common video encoders in existing TAL methods~\cite{Bai2020bcgnn,lin2019bmn,lin2018bsn,xu2020g}. Concretely, these works use two TSN networks, one with a ResNet50~\cite{he2016deep} backbone trained on RGB video frames, and the other with a BN-Inception backbone~\cite{ioffe2015batch} trained on optical flow. 
Other alternatives used as a video encoder for TAL include
two-stream I3D model~\cite{carreira2017quo} (see \cite{he2019rethinking,Zeng2019GraphCN}) and Pseudo-3D~\cite{qiu2017learning} (see \cite{Long2019GaussianTA}).

{In terms of optimization, 
a typical paradigm is two-staged:
first pre-training the video encoder and then, in a second stage,
training the TAL head of the model with the video encoder fixed. 
This is constrained by the inherent hardware budget derived from having a large per-video input size.
In particular,}
the video encoder is pre-trained using a cross-entropy loss for \textit{action recognition} on a large-scale video classification dataset such as Kinetics~\cite{zisserman2017kinetics,Zeng_2020_CVPR}.
An optional step is to further pre-train it 
on the foreground segments of the target TAL dataset
\cite{lin2019bmn,xu2020g,Rodriguez_2020_WACV,Mun_2020_CVPR}.
This brings a mismatch between training and inference for the video encoder,
which we call a task discrepancy problem.
More specifically, although trained to 
distinguish the content of different action classes,
the video encoder is less sensitive to action temporal boundaries and thus less effective for the TAL task. In fact, due to their inherent design, CNNs have limited localization capabilities~\cite{mallat2016understanding}, unless they are augmented with specialized localization-specific layers~\cite{liu2018intriguing}.
Additionally, the action classification task focuses only 
on the foreground content whilst ignoring per-class background
segments, including the transition between foreground and background.
In this paper we propose a novel low-fidelity  video encoder pre-training method to solve this limitation with existing TAL methods.

While current TAL literature mostly relies on pre-training through supervised learning, the rapid advancement of self-supervised learning makes it a promising alternative 
\cite{alwassel2020xdc,benaim2020speednet,miech20endtoend,shuffle_learn_eccv16,arrow_cvpr18}. 
Some works have focused on finding effective temporal-related pretext tasks, from frame ordering learnt through triplets of frames \cite{shuffle_learn_eccv16}, to sorting the frames of a sequence \cite{sorting_seq_iccv17}, distinguishing whether sequences are played forward or backwards \cite{arrow_cvpr18} or through playback speed-related pretext tasks~\cite{benaim2020speednet,playback_rate_cvpr20,wang2020self,temporal_transform_eccv20}. 
%
These methods exploit video-specific characteristics to force the network to focus on some sort of semantic content within video, inducing representations capturing long-term temporal semantic relations, but force invariance to or ignore the relative positioning of the snippets within the action instances. They are thus not suited for pre-training the video encoder of a TAL model. 

Very recent works~\cite{alwassel2020tsp,xu2020boundary} have exploited some of the aforementioned techniques for better pre-training of action localization models. 
For example, localization-tailored data augmentation and classification is adopted by \cite{xu2020boundary}. 
However, these works introduce a large amount of extra video data and additional stream networks, both of which are expensive in terms of memory and computation.
In contrast, our method aims to improve 
TAL modelling directly without the need for learning from extra training video data and using an expensive second network, nor relying on optical flow obtained at high computational cost.


\begin{figure*}
    \centering
    \vspace{-1cm}
    \includegraphics[trim={0.5cm 1.5cm 0.35cm 1.5cm}, width=0.95\textwidth,clip]{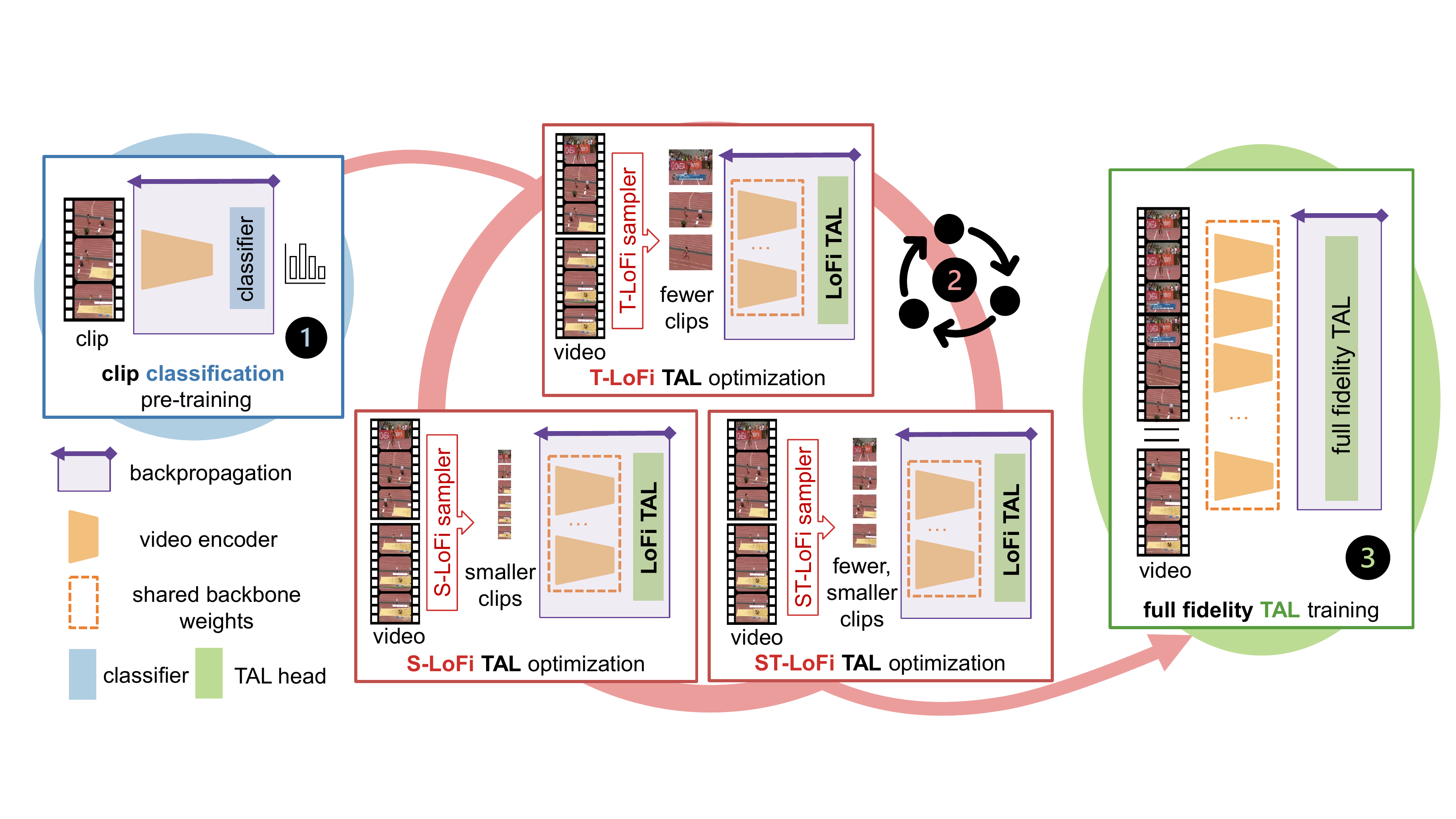}
    
    \vspace{-0.5cm}
    \caption{
    \textbf{Schematic overview of the proposed TAL model training procedure.}
    Three stages are involved during model training:
    (1) Pre-training the {\em video encoder} under action classification task's supervision on an auxiliary video dataset (\eg, Kinetics \cite{zisserman2017kinetics});
    (2) Low-fidelity (low mini-batch configuration in the spatial, temporal, or spatio-temporal resolution of training videos) {optimization} of the 
    {\em video encoder} together with the TAL head under TAL task's supervision on the target dataset;
    This is the key stage introduced in this paper for resolving the task discrepancy problem without memory overhead increase.
    (3) Training the {\em TAL head} in the full fidelity configuration under TAL task's supervision on the target dataset.
    }
    \label{fig:train_details}
\end{figure*}

\section{Method}

A TAL model takes as input 
a long untrimmed video 
with a varying number of frames. 
For design convenience,
{it is typical to represent a varying-length video by decomposing it} into 
a {fixed-length sequence of $L$ \textit{snippets}}. The definition of a snippet is the same as in action recognition, where first a number of consecutive frames (\eg, 64) is selected and then sub-sampled with stride $r$ (\eg, stride 8 to obtain 8-frame snippets).
To represent a snippet, one first applies a video encoder 
to extract frame-level feature vectors
and then averages them to obtain the snippet-level feature representation \cite{Buch2017SSTST, Escorcia2016DAPsDA, gao2017turn,lin2018bsn}.
%
%
The resulting snippet feature sequence is denoted as
$X\in \mathbb{R}^{C\times L}$, 
where $C$ is the feature dimension of each snippet, and $L$ is the number of snippets. 

In the training set, each video is associated with {its ground truth, consisting of} a set of 
action instance annotations
$\Psi$. {In particular, each action instance is represented as a segment,}
each including the start time, the end time, and the action class label.
%
The objective is to train a TAL model that can accurately localize all the target action instances in a given untrimmed video.
To that end, the model predicts a varying number of
%
action instances $\Phi$, {each comprised of the} predicted temporal boundaries, action class, and confidence score.
%
%

\subsection{Model Training Procedure}
Our training procedure 
consists of three stages as depicted in Figure \ref{fig:train_details}.
{\bf(1)} First, we pre-train the video encoder by action classification supervision on a large video dataset (\eg, Kinetics \cite{zisserman2017kinetics}).
{\bf(2)} Second, we conduct low-fidelity (\lofi{}) {optimization} 
of the video encoder with a \textit{TAL head} on the \textit{target dataset} (Sec.~\ref{sec:LF_design}).
This is under TAL supervision with the objective loss function derived from the ground-truth $\Psi$ and model prediction $\Phi$.
To this end, we propose to reduce the mini-batch configuration in terms of the spatial and temporal resolution of the input as otherwise end-to-end optimization cannot satisfy the hardware constraints. 
Crucially, by training on the target task and target dataset, 
the task discrepancy gap can be reduced.
{\bf(3)} Last, we freeze the already end-to-end {optimized} video encoder
and {train the TAL head of choice from scratch on the target dataset at full spatial and temporal resolutions}.
Note that in this setup, we cannot perform end-to-end optimization of the final TAL model, limited by the hardware memory constraint.
Next, we will detail the proposed~\lofi{} training.

\subsection{Low-Fidelity Training Configurations}
\label{sec:LF_design}

Formally, we define the {\em full fidelity} configuration of a mini-batch 
as 
\begin{equation}
    \Omega_f = L \times H \times W,
    \label{eq:full_cfg}
\end{equation}
%
where $L$ specifies the temporal resolution, and $H \times W$ refers to the 2D spatial resolution.
Under the full fidelity regime and with a certain memory size constraint, only the TAL head can be trained whilst leaving the video encoder frozen.
In order to enable the video encoder to be optimized
end-to-end together with the TAL head under the TAL task's supervision, 
we design four low-fidelity configurations for the mini-batch.
%

\paragraph{\bf (I) Spatial Low-Fidelity (\slofi{})}
In the first configuration, we lower the spatial resolution 
of the input videos by a factor of $r_s$ in both spatial dimensions as:
\begin{equation}
    \Omega_s = L \times (H/r_s) \times (W/r_s)
    \label{eq:s_lf}
\end{equation}
With smaller spatial feature maps, this could effectively reduce the memory consumption of the video encoder's feature maps which in turn creates space to enable the learning 
of the video encoder.

\paragraph{\bf (II) Temporal Low-Fidelity (\tlofi{})}
In the second configuration, we instead consider temporal resolution reduction in the following form:  
\begin{equation}
    \Omega_t = (L/r_t) \times H \times W,
    \label{eq:t_lf}
\end{equation}
where $r_t > 1$ is the scaling factor.
This corresponds to a smaller number of snippets
being taken as input by the TAL head, 
finally outputting less predictions in case the candidates per time location remains the same. 
{This reduces the memory demands of \textit{both} the video encoder and the temporal localization
{head}.}

\paragraph{\bf (III) Spatio-Temporal Low-Fidelity (\stlofi{})}
In the third configuration, we apply a (typically smaller) reduction in both 
temporal and spatial resolutions concurrently, formulated as:
\begin{equation}
    \Omega_{st} = (L/r_t)  \times (H/r_s) \times (W/r_s)
    \label{eq:st_lf}
\end{equation}
In this setup, memory saving can be shared between the
temporal and the spatial dimensions.

\paragraph{\bf (IV) Cyclic Low-Fidelity (\clofi{})}
Each of the above three \LF{} configurations is used in isolation.
To further explore their complementary benefits, we propose to apply {all of} them in a structured fashion. 
To that end,
{we take inspiration from the} recently proposed multi-grid training strategy \cite{wu2020multigrid}, originally designed for speeding-up action recognition training.
Specifically, we form a sampling grid with the three LoFi configurations
and cycle through them repeatedly.
We integrate both the short and long cycle strategies \cite{wu2020multigrid}.
In particular, the \textbf{long cycle} changes the values of $r_t$ and $r_s$ at the beginning of an epoch, and cycling through the configurations I to III after $c_l$ epochs. 
{Instead, the \textbf{short cycle} also cycles through configurations I to III, but changes $r_t$ and $r_s$ after $c_s$ batches instead.} 

It is important to note some practical differences with the original multri-grid training. {First,} contrary to the original multi-grid method, we shift the reduction in input resolution between the temporal and the spatial dimensions so that the {\em per-video} memory usage remains (approximately) constant. Thus, the batch size also remains constant throughout, which simplifies training. {Second, the full resolution setting (\ie, the full fidelity) is not used at any time during the video encoder training.} 

We compare the above four configurations in Section \ref{sec:exp_lf_design}.
By default, we use the \clofi{} configuration with the long cycle strategy (Long C-LoFi).

\subsection{An Instantiation of \lofi{}} \label{sec:detail}

\paragraph{Off-the-shelf TAL model:}
{Without loss of generality, in this study we adopt G-TAD~\cite{xu2020g}, a state-of-the-art TAL method, as our temporal localization module. However,  any other standard alternative, \eg~\cite{lin2019bmn}, can be adopted without any additional considerations
(see Table {A} in \textit{supplementary material}). 
}
Relying on graph convolutions \cite{kipf2017semi}, G-TAD is composed of a stack of GCNeXt  blocks to obtain context-aware features. In each GCNeXt block, there are two graph convolution streams to model two types of contextual information. One stream operates with temporal neighbors, and the other adaptively aggregates semantic neighbors in the snippet feature space. At the end of {the last} GCNeXt block, G-TAD extracts a set of sub-graphs based on pre-defined temporal anchors. With the sub-graph of interest alignment layer, SGAlign, it represents each sub-graph using a feature vector, which is further used as input to multiple fully-connected layers to predict the final action predictions.




To train G-TAD with our \lofi{},
we select one of the proposed low-fidelity variants and apply the original TAL loss function
to optimize both the video encoder and the TAL head. 
{We initialize G-TAD weights as in~\cite{xu2020g}.}


\paragraph{Implementation details:} 
We use ResNet-based TSM \cite{lin2019tsm} as the video encoder due to its good accuracy-cost trade-off and reasonable memory requirements compared to 3D-based alternatives.
For the full fidelity setting (Eq. (\ref{eq:full_cfg})), we follow the standard G-TAD protocol and represent each video with $L = 100$ snippets. The full spatial resolution is $H\times W = 224\times 224$.
We keep the other hyper-parameters (\eg, the number of GCNeXt layers) the same as in the default G-TAD configuration. However, the number of anchor proposals can be reduced when $L$ is less. Concretely, we enumerate all the possible combinations of start and end as the anchors, \eg, $\{(t_s,t_e)|~ 0< t_s<t_e<L; ~t_s,t_e \in \mathcal{N}; ~t_e-t_s < L\}$.

For \lofi{} training, we use an SGD optimizer. The batch size is 16 for all the training methods and input patterns. The weight decay is $10^{-4}$ and we set the momentum to 0, which is standard for fine-tuning~\cite{finetunning_iclr20}.
The learning rate is $0.1$, and it is decayed by 0.5 after every 5 epochs.


When we train G-TAD using the full fidelity setting, we keep the same training strategy as described in the original paper \cite{xu2020g}, except that we perform a learning rate search within the set \{0.0002, 0.0005, 0.001, 0.002, 0.005\}. We follow the rest of the common post-processing steps for TAL as specified in the original paper, including the application of soft-NMS \cite{softNMS} with a threshold of $0.84$. We select the top-$100$ predictions for the final evaluation. {We provide our code and script as part of the \textit{supplementary material}}. 

\paragraph{Hardware and software settings:}
We implemented our method using PyTorch 1.8 with CUDA 10.1. 
For \lofi{} training, we use 4 NVIDIA V100 GPUs, each with 32GB memory. 
Under this setting, the memory constraint is 128GB, which constitutes a mid-range computational budget. 
{In the \textit{supplementary material}, we further test a low-budget setting with a single V100 GPU {in Table B}.} 

\section{Experiments}

\subsection{Experimental setup}


\paragraph{\bf Datasets:}
We use \textbf{\em Kinetics400} \cite{zisserman2017kinetics} as the auxiliary video classification dataset for initial
pre-training of the video encoder.
For model performance evaluation,
we use two popular temporal action localization benchmarks.
(1) \textbf{\em ActivityNet-v1.3} \cite{caba2015activitynet} contains {20K} temporally annotated untrimmed videos with 200 action categories.
In the standard evaluation protocol, these videos are divided into the training/validation/testing  sets by the ratio of 2:1:1. 
(2) Human Action Clips and Segments (\textbf{\em HACS-v1.1})
\cite{zhao2019hacs}
is a recent temporal action localization dataset.
It contains 140K complete {action} segments from 50K videos
including 200 action categories (the same ones as ActivityNet-v1.3).


\noindent
\paragraph{\bf Evaluation metrics:} 
We adopt the mean Average Precision (mAP) rate at specified IoU thresholds as the main evaluation metrics. 
Following the standard evaluation setting, mAP values at a set of IoU thresholds, $\{0.5, 0.75, 0.95\}$ are reported,
as well as the average mAP over 10 different IoU thresholds $[0.5: 0.05:0.95]$.

\subsection{Pre-training Methodology Comparisons}

\paragraph{Setting:}

In this set of experiments, we directly compare different methodologies for pre-training the video encoder. To compare with our proposed \lofi{} {method}, we first consider the most widely used pre-training method that {optimizes} the video encoder through the action classification task on an auxiliary dataset (Kinetics400 in our case).
We denote this method as {\em Action Classification Pre-training} ({\bf ACP}).
To demonstrate the effect of video encoder pre-training,
we also take into account an ImageNet pre-trained video encoder, which uses no video data.
{We refer to this baseline as {\em Image Classification Pre-training} ({\bf ICP}).}
To adapt the video encoder to the target dataset, it can be further trained through an action classification task
on a clip version of the target TAL dataset, using each positive segment as a training instance.
We denote this variant as {\bf ACP+}.

In this experiment, 
we train our model using the cyclic low-fidelity (\clofi{}) configuration with the long cycle strategy. 
Other configurations will be evaluated 
in Section \ref{sec:exp_lf_design}.
%


\begin{table}[!h]
\centering
\caption{
\textbf{Comparing TAL results of different video encoder pre-training methods.} ACP: Action Classification Pre-training, {ICP: Image Classification Pre-training,}
ACP+: further fine-tuning ACP on the target dataset using a classification task using positive action segments.
}
\small
\vspace{0.1cm}
\begin{tabular}{c|ccc|>{\columncolor[gray]{0.9}}c|ccc|>{\columncolor[gray]{0.9}}c}
\toprule
Metric  & 0.5  &  0.75  & 0.95 & Average & 0.5  &  0.75  & 0.95 & Average\\
\midrule
Dataset & \multicolumn{4}{c|}{ActivityNet-v1.3} & \multicolumn{4}{c}{HACS-v1.1}
\\
\midrule
ACP & 49.64 & 34.16 & 7.68 & 33.59 & 35.68 & 22.79 & 6.51 & 23.00\\
ICP & 48.45 & 32.40 & 6.89 & 32.16 & 31.74 & 19.64 & 5.67 & 20.18\\
ACP+ & 49.87 & 34.58 & 7.85 & 33.84 & 36.31 & 22.96 & 6.62 & 23.31\\
\midrule
\bf \lofi{} (ours) 
&\bf 50.68 &\bf 35.16 &\bf 8.16 &\bf 34.49 &\bf 37.47 &\bf 24.36&\bf 7.08 &\bf 24.62
\\
\bottomrule
\end{tabular}
\label{tab:pre_train_comp}
\end{table}

\paragraph{\bf Results:}
The results of the different video encoder pre-training methods
are reported in Table \ref{tab:pre_train_comp}.
We make the following observations.
{\bf (1)} Without pre-training on a related auxiliary dataset,
the model performance could be significantly degraded {(see row 1 \vs{} row 2)}.
This indicates the significance of the video encoder and its pre-training on large, relevant video data.
%
{\bf (2)}
With action classification based pre-training on the target video data, the performance indeed improves to some degree {(see row 1 \vs{} row 3)}.
This means that addressing the data distribution shift between
the auxiliary dataset and the target dataset is {essential}.
{\bf (3)}
Importantly, the biggest performance gains come from the proposed \lofi{} method, which instead solves the task discrepancy issue by {optimizing} the video encoder with the TAL task {(see the last row)}. This indicates that although auxiliary video data is similar to the target data in distribution, the task-level differences would still pose obstacles harming the model performance. This confirms the motivation and hypothesis of this study.
Once this obstacle is properly tackled with  our low-fidelity pre-training, 
more significant performance gains can then be rewarded.
Overall, this verifies the efficacy of the proposed method 
in pre-training the video encoder.


\subsection{\lofi{} Configuration Comparisons}
\label{sec:exp_lf_design}

\paragraph{Setting:}
We investigate different \lofi{} configurations.
We use the setting as:
$r_t=4$ for \tlofi{},
$r_s=2$ for \slofi{} (note that this is applied in both spatial dimensions, thus being comparable to \tlofi{}),
$r_s=\sqrt{2}, r_t=2$ for \stlofi{},
and {$c_s=16$, $c_l=1$ for \clofi{}'s short and long cycle strategies, respectively.}
We include the standard pre-training strategy (\ie, ACP) to provide a baseline and facilitate comparisons.


\begin{table*}[!h]
\centering
\caption{\textbf{Comparing different low-fidelity configurations}. Dataset: ActivityNet-v1.3 and HACS-v1.1. 
ACP: Action Classification Pre-training.
TR: Temporal Resolution;
SR: Spatial Resolution.
}
\small
\vspace{0.1cm}
\begin{tabular}
{l|cc|ccc|>{\columncolor[gray]{0.9}}c|ccc|>{\columncolor[gray]{0.9}}c}
\toprule
&&&\multicolumn{4}{c|}{ActivityNet-v1.3}& \multicolumn{4}{c}{HACS-v1.1} \\ 
Configuration & TR & SR 
& 0.5  &  0.75  & 0.95 & Avg & 0.5  &  0.75  & 0.95 & Avg \\
\midrule
ACP & - & 224$\times$224 & 49.64 & 34.16 & 7.68 & 33.59 
& 35.68 & 22.79 & 6.51 & 23.00
\\
\midrule
\slofi{} & 100 & 112$\times$112 
& 50.47 & 34.71 & 7.57 & 34.12 & 37.16 & 24.23 & 6.84 & 24.25
\\
\tlofi{} & 25 & 224$\times$224 
& 50.28 & 35.21 & 8.09 & 34.32 &  37.30 & 24.18 & 7.07 & 24.36
\\
\stlofi{} & 50 & 158$\times$158 
& 50.36 & 34.79 &  7.59 & 34.12  & 37.13 & 24.27 & 7.08 & 24.36
\\
\midrule
Short \clofi{} & \multicolumn{2}{c|}{Batch-level Cycle} 
& 50.57 & 35.12 & 8.14 & 34.38 & 37.63 & \bf 24.45 & 6.95 & 24.60 \\
Long \clofi{} & \multicolumn{2}{c|}{Epoch-level Cycle} 
&  \bf 50.68 &\bf 35.16 &\bf 8.16 &\bf 34.49 
& \bf 37.78 & 24.40 & \bf 7.29 & \bf 24.64 \\
\bottomrule
\end{tabular}
\label{tab:lf_design}
\end{table*}

\paragraph{Results:}
The results for ActivityNet-v1.3 and HACS-v1.1 are shown in Table \ref{tab:lf_design}.
We provide the following observations.
{\bf (1)}
Each of our proposed \LF{} configurations can improve the 
{video encoder, thus suggesting that spatial and temporal dimensions 
 are both good selections} for low-fidelity manipulation.
{\bf (2)}
Integrating our \LF{} configurations into a more advanced
cyclic training strategy produces some further, although moderate, improvement. In particular, the long cycle leads to the best overall performance for both datasets.




\begin{table}[!h]
\centering
\caption{\textbf{Comparing TAL results with state-of-the-art methods on ActivityNet-v1.3 validation set}. 
``*'' indicates RGB-only Kinetics400 pre-trained TSM video encoder without fine-tuning. 
O.F.: Optical Flow.
R18/50: ResNet-18/50.
}
\small
\vspace{0.1cm}
\begin{tabular}
{l|ccc|ccc|>{\columncolor[gray]{0.9}}c}
\toprule
Method & O.F. & Arch. & \#Par & 0.5  &  0.75  & 0.95 & Average \\
\midrule
SCC \cite{heilbron2017scc}  & \xmark  & C3D & 79M & 40.00 & 17.90  & 4.70   & 21.70  \\
CDC \cite{shou2017cdc} & \xmark & C3D & 79M & 45.30 & 26.00 & 0.20 & 23.80 \\
R-C3D \cite{xu2017r} & \xmark & C3D & 79M & 26.80 & - & - & - \\
BSN \cite{lin2018bsn}& \cmark  & R50 & 23M& 46.45  & 29.96 & 8.02  & 30.03  \\
P-GCN \cite {Zeng2019GraphCN}& \cmark & I3D & 25M
&48.26 &33.16 &3.27 &31.11  \\
BMN \cite{lin2019bmn} & \cmark & R50 & 23M & { 50.07} & {34.78} & { 8.29} & { 33.85}  \\
{BC-GNN}~\cite{Bai2020bcgnn} & \cmark & R50 & 23M & {50.56} & {34.75} & {\textbf{9.37}} & { 34.26} \\

{G-TAD}~\cite{xu2020g}& \cmark & R50 & 23M & {50.36} & {34.60} & {{9.02}} & { 34.09}  \\ \midrule
{G-TAD} baseline* & \xmark & R18 & 12M & {49.64} & {34.16} & {7.68} & { 33.59} \\
{G-TAD}+ \bf \lofi{}& \xmark  & R18 & 12M & \bf 50.68 & \textbf{35.16} & {8.16} & {\bf 34.49} \\
\bottomrule
\end{tabular}
\label{tab:sota_anet}
\end{table}

\begin{table}[!h]
\centering
\caption{\textbf{Comparing TAL results with state-of-the-art methods on HACS-v1.1 validation set}. 
``*'' indicates RGB-only Kinetics400 pre-trained TSM video encoder without fine-tuning. 
O.F.: Optical Flow.
R18/50: ResNet-18/50. 2S.: 2-stream.
}
\small
\vspace{0.1cm}
\begin{tabular}
{l|ccc|ccc|>{\columncolor[gray]{0.9}}c}
\toprule
Method & O.F. & Arch & \#Par & 0.5  &  0.75  & 0.95 & Average \\
\midrule
SSN \cite{Zhao2017TemporalAD} & \cmark & 2S & 12M & 28.82& 18.80 & 5.32 & 18.97 \\ \midrule
{G-TAD} baseline* & \xmark & R18 & 12M & 35.68 & 22.79 & 6.51 & 23.00 \\
{G-TAD} + \bf \lofi{}& \xmark  & R18 & 12M & \bf 37.78 & \bf 24.40 & \bf 7.29 & \bf 24.64 \\
\bottomrule
\end{tabular}
\label{tab:sota_hacs}
\end{table}

\subsection{Comparison with State-of-the-Art}

\paragraph{Setting:}
Following video encoder pre-training evaluation, we further conduct a system-level performance comparison
with previous state-of-the-art methods.
Whilst ResNet50-based (R50) methods are a common backbone choice for the video encoder, we still use a more lightweight ResNet18 (R18) for our method 
due to its higher efficiency in computation and memory.
Furthermore, compared to the popular architecture design that uses two streams (one for RGB and one for optical flow),
our method only uses RGB frames, avoiding the excessive
costs incurred from computing optical flow and running a second forward pass. 

\paragraph{Results:}
The results of our method are compared with existing alternatives in Table \ref{tab:sota_anet} for ActivityNet-v1.3 and 
Table \ref{tab:sota_hacs} for HACS-v1.1.
It is evident that our method can achieve the best 
performance among all the competitors, despite
the single stream input modality and a much lighter video encoder backbone.
This clearly demonstrates that tackling the pre-training of the video encoder is of particular importance for TAL, and that existing efforts towards improving the TAL head model have neglected it as a key model component.
On ActivityNet-v1.3, it is encouraging to see that with a stronger pre-trained video encoder using our method and a shallower architecture,
optical flow can be favourably eliminated 
without performance sacrifice (actually even enjoying better performance).
This result is substantial, since the study of means to get rid of optical flow for more efficient action analysis  is itself an important research problem \cite{crasto2019mars}.

\subsection{Using Different Video Encoders}

\paragraph{Setting:}
Our \lofi{} method can be used in combination with different video encoders, as long as the backbone is end-to-end trainable.
We compare the performance gains of our default encoder, \ie ResNet18-based TSM, with a 3D-based alternative, \ie an 18-layer R(2+1)D encoder \cite{tran2018closer}, on ActivityNet-v1.3. Note that the default spatial resolution for R(2+1)D, as defined by the authors, is $112\times 112$ pixels. We thus use \tlofi{} and maintain the default spatial resolution for both networks.

\begin{table}[!h]
\centering
\caption{\textbf{Ablation on different video encoders.} The performance improvement from using our \LF{} pre-training does not depend on the video encoder's backbone. Dataset: ActivityNet-v1.3.
TAL head: G-TAD.
R18: ResNet18.
}
\small
\vspace{0.1cm}
\begin{tabular}
{l|ccc|>{\columncolor[gray]{0.9}}l|ccc|>{\columncolor[gray]{0.9}}l}
\toprule
&\multicolumn{4}{c|}{TSM-R18 backbone}& \multicolumn{4}{c}{R(2+1)D-R18 backbone} \\ 
Method & 0.5  &  0.75  & 0.95 & Avg & 0.5  &  0.75  & 0.95 & Avg \\
\midrule
{G-TAD} + ACP & 49.64 & 34.16 & 7.68 & 33.59
& 49.65 & 34.11 & \textbf{8.66} & 33.55 
\\
{G-TAD} + \tlofi{} & \textbf{50.28} & \textbf{35.21} & \textbf{8.09} & \textbf{34.32}
& \textbf{49.84} & \textbf{34.73} & 8.64 & \textbf{34.21}
\\ \midrule
\multicolumn{1}{c|}{{\textit{Gain}}} & +0.64 & +1.05 & +0.41 & +0.73
& +0.19 & +0.62 & -0.02 & +0.66 \\
\bottomrule
\end{tabular}
\label{tab:backbone}
\end{table}

\paragraph{Results:}
The results are summarized in Table \ref{tab:backbone}.
We observe that with either TSM-R18 or R(2+1)D as video encoder,
our method can similarly and consistently improve the performance of
the state-of-the-art G-TAD method. This verifies that our \lofi{} method is generally effective and useful
in training TAL models.




\subsection{Scaling up to ResNet-34 and ResNet-50}

\paragraph{Setting:}
{To evaluate the generalizability of our \lofi{},
we further evaluate two deeper video encoders: 
ResNet34/50-based TSM.
We use the long cyclic low-fidelity configuration.
Limited by the GPU memory size, we reduce the batch size
accordingly whilst keeping all other hyper-parameters and the training protocol unchanged.
This experiment is conducted on ActivityNet-v1.3 in comparison  with
the standard action classification pre-training (ACP) method as baseline.}

\begin{table}[!h]
\centering
\caption{\textbf{Ablation on different depths of TSM video encoders.} 
The performance improvement from using our \LF{} pre-training does not depend on the video encoder's backbone. Dataset: ActivityNet-v1.3.
TAL head: G-TAD.
R18: ResNet18; R34: ResNet34; R50: ResNet50.
}
\small
\vspace{0.1cm}
\begin{tabular}
{l|l|ccc|>{\columncolor[gray]{0.9}}l} 
\toprule
Encoder & Method & 0.5  &  0.75  & 0.95 & Avg. (\em Gain) \\
\midrule
TSM-R18 & ACP  & 49.64 & 34.16 & 7.68 & 33.59  \\
TSM-R18 & \bf Long C-LoFi  & \textbf{50.68} & \textbf{35.16} & \textbf{8.16} & \textbf{34.49} (+0.90)   \\
\midrule
TSM-R34 & ACP  & 50.16 & 34.35 & 8.31 & 33.90  \\
TSM-R34 & \bf Long C-LoFi  & \textbf{50.79} & \textbf{35.39} & \textbf{8.38} & \textbf{34.74} (+0.84)   \\
\midrule
TSM-R50 & ACP  & 50.32 & 35.07 & 8.02 & 34.26  \\
TSM-R50 & \bf Long C-LoFi  & \textbf{50.91} & \textbf{35.86} & \textbf{8.79} & \textbf{34.96} (+0.70)   \\
\bottomrule
\end{tabular}
\label{tab:scale}
\end{table}

{\bf Results:} 
{Table~\ref{tab:scale} shows that
whilst the gains with deeper encoders are slightly reduced,
\lofi{} can still consistently improve the results. This suggests good generalizability properties across varying-capacity networks.}

\subsection{Comparison to Self-Supervised Learning}

\paragraph{Setting:}
We compare our method to two representative video self-supervised learning methods: Arrow of Time~\cite{arrow_cvpr18} and SpeedNet~\cite{benaim2020speednet}.
{For both competitors, TSM-R18 is used as video encoder, and pre-trained on the Kinetics400 dataset. We use ActivityNet-v1.3 dataset for this experiment.}

\begin{wraptable}{r}{7.5cm}

\vspace{-0.7cm}
\centering
\caption{\textbf{Comparison to self-supervised learning methods.}
We use TSM-R18 as video encoder. Dataset: ActivityNet.
}
\small
\begin{tabular}{l|ccc|>{\columncolor[gray]{0.9}}l}
\toprule
Method   & 0.5  &  0.75  & 0.95 & Average\\
\midrule
ACP   & 49.64 & 34.16 & 7.68 & 33.59 \\ \midrule
Arrow \cite{arrow_cvpr18}  & 44.14 & 28.87 & 5.90 & 28.82 \\
Speed \cite{benaim2020speednet}  & 44.50 & 29.52 & 6.14 & 29.39 \\ \midrule
ACP+Arrow & 49.79 & 34.48 & 7.70 & 33.72 \\
ACP+Speed & 49.84 & 34.11 & 7.50 & 33.75 \\ \midrule
\bf \lofi{} &\bf 50.68 &\bf 35.16 &\bf 8.16 &\bf 34.49 \\
\bottomrule
\end{tabular}
\label{tab:ssl}
\vspace{-0.5cm}
\end{wraptable}

\paragraph{Results:} 
From Table~\ref{tab:ssl}, we see that both SSL methods without action classification based supervision are clearly inferior for the TAL task (see row 2 and 3).
When combined with the standard action classification pre-training (ACP), 
only slight performance gains can be achieved despite doubling the computational cost in video encoding (see row 4 and 5).
This suggests marginal complementary benefit.
Overall, this result indicates that the proposed \lofi{} optimization is superior over recent SSL alternatives in pre-training TAL's video encoder. 


\section{Conclusion}
In this work we have presented a simple and effective low-fidelity (\lofi{}) video encoder optimization method for achieving more effective TAL models. 
This is motivated by an observation that in existing TAL methods,
the video encoder is merely pre-trained by action classification supervision on short video clips, lacking desired optimization {\wrt} the temporal localization supervision {on the target dataset}.
Indeed, {joint} optimization itself is not novel. 
However, this is non-trivial to conduct for training a TAL model, due to large per-video size that would easily overwhelm the GPU memory, rendering it infeasible in practice.
To overcome this obstacle, we propose to reduce the mini-batch construction configurations in the temporal {and/or} spatial dimensions of training videos so that end-to-end optimization becomes operable 
under the same memory condition.
%
Extensive experiments demonstrate that our method can clearly improve the performance of existing off-the-shelf TAL models, yielding new state-of-the-art performance even with only RGB modality as input and a more lightweight backbone based single-stream video encoder on two representative TAL benchmarks.

\section{Ethical considerations and broader impact}
\label{sec:ethical}

Any inherent bias present in the training data is likely to be captured by the learning algorithm given its data-driven nature. Deploying of the model into real-world scenarios should thus take that aspect into account. Furthermore, the method discussed in this paper focuses on a fundamental problem and as such its potential applications are hard to predict. However, to the best of our knowledge, there are currently no applications of this technology that raise ethical issues. Finally, biases based on gender, race or sexuality are unlikely, although not impossible.

\section{Acknowledgements}
This work was supported by the King Abdullah University of Science and Technology (KAUST) Office of Sponsored Research through the Visual Computing Center (VCC) funding.







\bibliographystyle{ieee_fullname}
\bibliography{egbib}




\appendix

\section{Additional Experiments}

\subsection{LoFi optimization works with other TAL models}

Our \lofi{} optimization is a model-agnostic optimization method. 
In the main paper, we have evaluated \lofi{} using G-TAD~\cite{xu2020g} as the TAL head.
To evaluate its generality, 
we further test Boundary Matching Network (BMN)~\cite{lin2019bmn}
which generates temporal action proposals for an existing action
classifier \cite{zhao2017cuhk} to predict the final results.


\paragraph{Settings:} 
We compare our \tlofi{}
with the Action Classification Pre-training (ACP) baseline that optimizes the video encoder through the action classification task on an auxiliary dataset (Kinetics400 \cite{zisserman2017kinetics} in our case).
We use the same hyper-parameter setting as in G-TAD case,
with ResNet-18 as the video encoder backbone.
We use publicly available BMN code
\footnote{https://github.com/JJBOY/BMN-Boundary-Matching-Network}. 
As the performance metrics, we follow the standard AR@$k$ (the average recall of the top-$k$ predictions) and AUC (the area under the recall curve).

\begin{table}[!h]
\centering
\caption{
\textbf{
Comparing TAL results of BMN on ActivityNet-1.3 validation set
when using two different video encoder optimization methods (ACP vs. \lofi{}).
} 
ACP: Action Classification Pre-training.
}
\small
\vspace{0.1cm}
\begin{tabular}{c|cccc|>{\columncolor[gray]{0.9}}c}
\toprule
Metric  & AR@1 & AR@5& AR@10&AR@100&  AUC \\
\midrule
ACP 
& $33.29$ & $48.90$ & $56.20$ & $74.88$ & $66.81$ \\
\bf \lofi{} (ours) 
& $\bf 33.71$ & $\bf49.41$ & $\bf56.81$ & $\bf75.58$ &  $\bf67.49$ 
\\
\textit{gain} & $+0.32$ & $+0.51$ & $+0.61$ & $+0.70$ & $+0.58$ \\
\bottomrule
\end{tabular}
\label{tab:bmn}
\end{table}

\paragraph{Results:} The results are reported in Table \ref{tab:bmn}. We observe that under all the evaluation metrics, our LoFi method consistently improves BMN's performance compared to ACP.
Together with the performance gain for G-TAD,
this verifies that LoFi is generally effective in improving TAL models.

\subsection{Performance-Hardware Budget Trade-off}
In the main paper, we have conducted the experiments with a fixed computational budget of 4 V100 GPUs. 
To test our \lofi{} in varied hardware conditions,
we further compare the model performance trade-off under two different budget cases.

\paragraph{Setting:} 
For a {\em low-budget} case, we define a configuration so that the whole training procedure can fit in a single V100 GPU (32GB)\footnote{Admittedly, a V100 GPU is still a high-end GPU, having 32GB of memory. The current on-demand hourly rate at AWS is 3.06USD.}. In this setting, the temporal resolution needs to be lowered to $25$ snippets and the spatial resolution to $112\times 112$ pixels.
For a {\em middle-budget} case, we consider 
4 V100 GPUs (128G). Under this setting, we further introduce a novel 
configuration (termed as \tlofi{}): 
using a lower spatial resolution ($112^2$) in exchange for a larger video backbone, ResNet-50.

\begin{table}[!h]
\centering
\caption{\textbf{Trade-off analysis between performance and budget (GPU memory)
on ActivityNet-1.3 validation set}. 
TR: Temporal Resolution.
SR: Spatial Resolution.
R18/50: ResNet-18/50.
}
\small
\vspace{0.1cm}
\begin{tabular}{l|cc|ccc|>{\columncolor[gray]{0.9}}c}
\toprule
TR/SR & Arch. & GPU & 0.5  &  0.75  & 0.95 & Average \\
\midrule
ACP & R18 & -- & 49.64 & 34.16 & 7.68 & 33.59 \\
\midrule
25/$112^2$ & R18 & 32G & 50.01 & 34.46 & \textbf{8.38} & 33.99 \\
25/$224^2$ & R18 & 128G & \textbf{50.28} & 35.21 & 8.09 & \textbf{34.32}  \\
25/$112^2$ & R50 & 128G & 50.07 & \textbf{35.31} & 8.03 & 34.23  \\
\bottomrule
\end{tabular}
\label{tab:budget}
\end{table}

\paragraph{Results:}
The performances of our method under different budget settings are compared in Table~\ref{tab:budget}.
We have these observations:
(1) It is seen that our method can still outperform the standard action classification pre-training (ACP) baseline under the low computational budget setting (see the first two rows).
(2) For the middle-budget case, 
it is found that lowering the spatial resolution for using a deeper video encoder (ResNet-50) leads to a slightly worse trade-off in Average-mAP;
Besides, 
using a ResNet18-based video encoder offers a clear efficiency advantage
at inference.

\subsection{Effectiveness of Joint Optimization}
While our proposed method effectively closes the domain and task gaps for TAL, it still has to sacrifice input spatial and/or temporal resolution. Thus, although arguably less damaging in terms of the final performance, there still exists a gap between train-time and test-time settings (low-fidelity \vs{} full-fidelity). 
In this section, we set both train-time and test-time to low-fidelity settings to evaluate the exact benefit from joint optimization of video encoder and TAL head.

\paragraph{Setting:}
We train G-TAD using 25 snippets (\ie, T-LoFi, $r_t=4$) in the following two settings: 1) using a video encoder pre-trained on Kinetics400 (\ie, ACP baseline) and 2) using end-to-end training.

\begin{table}[!h]
\centering
\caption{\textbf{
Comparing the video encoders pre-trained on Kinetics400 and end-to-end  training.} We use the \LF{} setting 
(25 snippets, $224^2$ res.)
} 
\small
\vspace{0.1cm}
\begin{tabular}
{l|c|ccc|l}
\toprule
Method & Dataset & 0.5  &  0.75  & 0.95 & Avg. (Gain) \\
\midrule
pre-trained & ActivityNet-v1.3 & 45.75 & 32.05 & 4.80 & 31.02  \\
end-to-end & ActivityNet-v1.3 & \textbf{47.52} & \textbf{33.30} & \textbf{5.31} & \textbf{32.21} (+1.19) \\ \midrule
pre-trained & HACS-1.1 & 29.28 & 17.95 & 4.05 & 18.49  \\
end-to-end & HACS-1.1 & \textbf{31.08} & \textbf{19.74} & \textbf{4.29} & \textbf{19.94}  (+1.45) \\
\bottomrule
\end{tabular}
\label{tab:pf}
\end{table}

\paragraph{Results:} The resulting performances are compared for both ActivityNet-v1.3 and HACS-1.1 in Table~\ref{tab:pf}. We can see that, while the absolute performance is significantly lower due to the lack of temporal resolution, the accuracy improves significantly on both datasets, namely 1.19 and 1.45 in terms of Average-mAP.
We also note that higher gains are achieved on HACS-1.1 than on ActivityNet-v1.3. 
We hypothesize that this is due to the more considerable amount of training data available on HACS-1.1, resulting in more benefits from our \lofi{} in video encoder pre-training.

\subsection{Training Stability}
In this section, we show our model robustness via the standard derivation of different evaluation metrics. We emphasize that the primary evaluation metric of TAL model is the average of mAP under the ten difference IoU thresholds, and mAP is the mean of Average-Precision over $k=200$ action classes. Thus, average mAP has a naturally robust property and, more importantly, \emph{a relatively small gain is still significant}.

\paragraph{Setting:}
With G-TAD as the TAL model, 
we experiment each optimization method (ACP, \lofi{}) for 10 trials
to test their stability.
We report the average result with the standard derivation on ActivityNet-1.3 dataset. We use ResNet18 as the video encoder backbone.

\begin{table}[!h]
\centering
\caption{
\textbf{Comparing TAL results of different video encoder pre-training methods.} ACP: Action Classification Pre-training, 
}
\small
\vspace{0.1cm}
\begin{tabular}{c|ccc|>{\columncolor[gray]{0.9}}c}
\toprule
Metric  & 0.5  &  0.75  & 0.95 & Average \\
\midrule
ACP & $49.64\pm 0.09$ & $34.16\pm 0.05$ & $7.68\pm 0.17$ & $33.59\pm0.04$ \\
\bf \lofi{} (ours) 
&\bf $50.68\pm 0.12$ &\bf $35.16\pm 0.07$ &\bf $8.16\pm 0.16$ &\bf $34.49\pm 0.03$
\\
\bottomrule
\end{tabular}
\label{tab:robust}
\end{table}

\paragraph{Results:}
The results in Table~\ref{tab:robust} show that both methods are
stable with some slight advantage of our \lofi{} in Average-mAP.


\section{Further Discussion}
\subsection{Our method's limitations}

A main limitation imposed by our proposed \lofi{} optimization method 
is that an extra training stage is introduced, which increases the model training complexity. However, due to the nature of low mini-batch configurations, the computational cost is still tolerable.


\subsection{Social Impact}
The research presented in the paper has a potential to positively contribute to a number of practical applications where understanding human's actions and events in video is critical, for example, pedestrian safety in automotive settings, patient monitoring in hospitals and elderly care homes. However, there is also a risk for the technology to be used for nefarious purposes, for example, in the area of unauthorized and immoral surveillance particularly by autocratic regimes. 
For partial mitigation, we commit to not authorize our technology to be used by any government bodies with such predictable risks.






\end{document}